\documentclass[graybox]{svmult}

\usepackage{type1cm}        
 \usepackage{ulem}                            
\usepackage{makeidx}         
\usepackage{graphicx}        
\usepackage{multicol}        
\usepackage[bottom]{footmisc}

\usepackage{newtxtext}       %
\usepackage[varvw]{newtxmath}       

\usepackage[capitalise]{cleveref}

\definecolor{bostonuniversityred}{rgb}{0.8, 0.0, 0.0}

\usepackage{tikz}
\usetikzlibrary{matrix}
\usetikzlibrary{shapes,patterns,snakes,arrows}
\usetikzlibrary{positioning,fit,calc}
\usetikzlibrary{graphs}

\tikzset{%
  every neuron/.style={
    circle,
    draw,
    minimum size=0.6cm
  },
  every input neuron/.style={
    circle,
    draw,
    minimum size=0.6cm,
    fill=green!50
  },
  every output neuron/.style={
    circle,
    draw,
    minimum size=0.6cm,
    fill=orange!30
  },
  every hidden neuron/.style={
    circle,
    draw,
    minimum size=0.6cm,
    fill=blue!40
  },
  neuron missing/.style={
    draw=none, 
    scale=1.5,
    text height=0.3cm,
    execute at begin node=\color{black}$\vdots$
  },
}

\makeindex             

\begin{document}

\title*{Model Parallel Training and Transfer Learning for Convolutional Neural Networks by Domain Decomposition}
\titlerunning{Model Parallel Training and Transfer Learning for CNNs by Domain Decomposition}

\author{Axel Klawonn, Martin Lanser, and Janine Weber}
\institute{Axel Klawonn, Martin Lanser, Janine Weber \at Department of Mathematics and Computer Science, University of Cologne, Weyertal 86-90,\\ 50931 K\"oln, Germany, \url{https://www.numerik.uni-koeln.de}\\
Center for Data and Simulation Science, University of Cologne, 50923 K\"oln, Germany, \url{https://www.cds.uni-koeln.de}\\  \email{\{axel.klawonn,martin.lanser,janine.weber\}@uni-koeln.de}}
%
%
\maketitle

\abstract{Deep convolutional neural networks (CNNs) have been shown to be very successful in a wide range of image processing applications. However, due to their increasing number of model parameters and an increasing availability of large amounts of training data, parallelization strategies to efficiently train complex CNNs are necessary. 
In previous work by the authors, a novel model parallel CNN architecture was proposed which is loosely inspired by domain decomposition. In particular, the novel network architecture is based on a decomposition of the input data into smaller subimages. For each of these subimages, local CNNs with a proportionally smaller number of parameters are trained in parallel and the resulting local classifications are then aggregated in a second step by a dense feedforward neural network (DNN). In the present work, we compare the resulting CNN-DNN architecture to less costly alternatives to combine the local classifications into a final, global decision. Additionally, we investigate the performance of the CNN-DNN trained as one coherent model as well as using a transfer learning strategy, where the parameters of the pre-trained local CNNs are used as initial values for a subsequently trained global coherent CNN-DNN model. }

\section{Introduction}
\label{sec:intro}

Convolutional neural networks (CNNs)~\cite{lecun:1989:CNN} have been shown to be tremendously successful in processing image data or, more general, data with a grid-like structure. However, with increasing numbers of model parameters and increasing availability of large amounts of training data, parallelization approaches for a time- and memory-efficient training process have become increasingly important; see also~\cite{ben2019demystifying} for an overview. 
In general, most parallelization approaches can be categorized into model or data parallel methods~\cite{ben2019demystifying}. In data parallel approaches, different cores or processors of a parallel machine obtain local copies of the underlying deep learning model which are trained with local subsets of training data points. Usually, the locally trained models are then aggregated once or iteratively after a fixed number of epochs to obtain a final, global model. 
In model parallel approaches, not the training data but the neural network model itself is distributed to different cores or processors of a CPU or, typically, a GPU. Depending on the decomposition of the network architecture, the total global model then needs to be composed from the locally trained network parameters either once, at the end of the training, or frequently, given that in neural networks, one layer usually needs the output of the previous layer. 

Generally speaking, many model parallel training approaches can be interpreted as domain decomposition methods (DDMs)~\cite{toselli}; see~\cite{KLW:DD_ML_survey:2023} for a survey of existing approaches based on the combination of machine learning and DDMs. 
In~\cite{KLW:DNN-CNN:2023}, a novel model parallel training strategy for CNNs applied to different image classification problems has been presented. This training strategy is based on a decomposition of the input images into smaller subimages and hence, proportionally smaller CNNs operating exclusively on the subimages are trained in parallel. In particular, the training of the local CNNs does not require any communication between the different local models. Subsequently, a dense feedforward neural network (DNN) is trained that evaluates the resulting local classification probability distributions into a final global decision.  
Due to the divide-and-conquer character as well as the implementation of a global coupling between the different local CNN models, the described method can be loosely interpreted as a domain decomposition approach. 

In this paper, we extend our previous work from~\cite{KLW:DNN-CNN:2023} by several comparisons. First, we provide further comparative results of the CNN-DNN model from~\cite{KLW:DNN-CNN:2023} with computationally less costly alternatives to combine the local CNN classifications into a final, global decision. Second, we present classification accuracies for training the CNN-DNN model from~\cite{KLW:DNN-CNN:2023} as one cohesive model architecture. 
Finally, we additionally consider the idea of transfer learning such that the network parameters of the locally trained CNN models are used as initial values for a subsequently trained global coherent CNN-DNN model.

\section{Training strategies}
\label{sec:train_strat}

In this section, we briefly describe the parallel CNN-DNN model architecture as introduced in~\cite{KLW:DNN-CNN:2023} as well as its extended variants and modifications for transfer learning which are considered in this paper for the first time. 
Let us note that the idea of decomposing a CNN into smaller subnetworks within the context of preconditioning and transfer learning has also been considered in~\cite{GuCai:2022:dd_transfer}. However, the globally trained CNN model in~\cite{GuCai:2022:dd_transfer} is different from our globally trained network architecture.

\subsection{Parallel CNN-DNN model architecture}
\label{sec:CNN-DNN}

\begin{figure}
\centering
\includegraphics[width=0.9\textwidth]{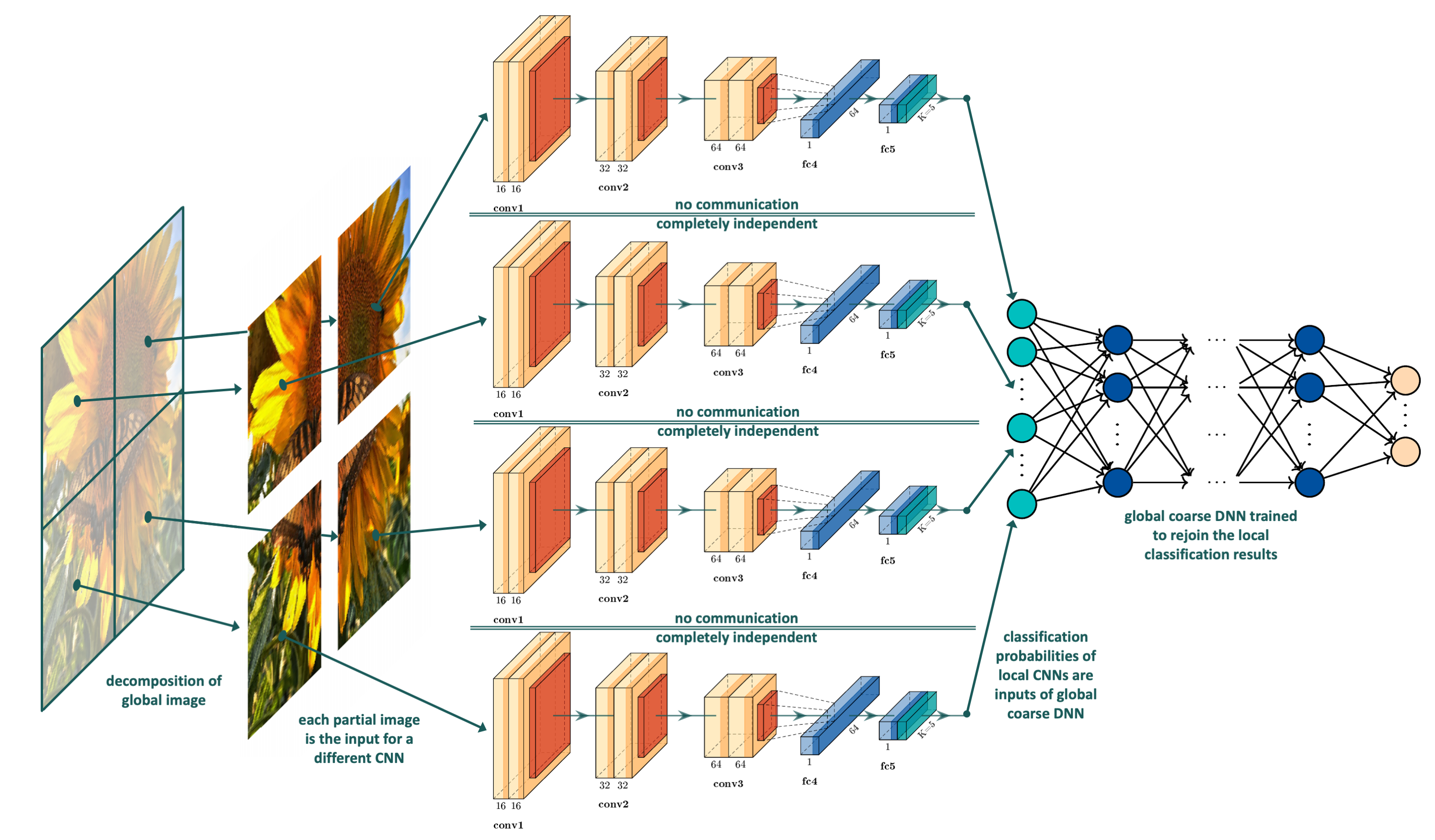}
\caption{Visualization of the CNN-DNN network architecture. \textbf{Left:} The original image is decomposed into $N=4$ nonoverlapping subimages. \textbf{Middle:} The $N=4$ subimages are used as input data for $N$ independent, \textit{local} CNNs. \textbf{Right:} The probability values of the local CNNs are used as input data for a DNN. The DNN is trained to make a final classification for the decomposed image by weighting the local probability distributions.
Figure taken from~\cite[Fig. 4]{KLW:DNN-CNN:2023}.}
\label{fig:dd_cnn}
\end{figure}

As presented in~\cite{KLW:DNN-CNN:2023}, we consider a hybrid CNN-DNN neural network architecture which naturally supports a model parallel training strategy. 
As a starting point, we assume that we have a classic CNN model that takes as input data a two-dimensional pixel image with $H\times W$ pixels and outputs a probability distribution with respect to $K \in \mathbb{N}$ classes. In order to define our CNN-DNN model, we now decompose the input data in form of images into a finite number of $N \in \mathbb{N}$ smaller subimages. Note that for colored input images with 3 channels of $H\times W$ pixels, we exclusively decompose the images in the first two dimensions, the height and the width, but not in the third dimension. Hence, each image is decomposed into $N$ subimages with height $H_i$ and width $W_i, \ i=1, \ldots, N$. Then, for each of these subimages, we construct corresponding subnetworks,  that is, local CNNs that only operate on certain subimages of all input images. Let us note that, in this paper, due to space limitations, we exclusively consider decompositions of the input images into rectangular subimages without overlap. We refer to this type of decomposition as type A decomposition; refer also to~\cite[Sect. 3.1]{KLW:DNN-CNN:2023} for more general and overlapping decompositions of the input images. Analogously, the described decomposition of the input data can also be generalized to three-dimensional data, that is, voxel data, for example, in form of computed tomography (CT) scans. In that case, the considered CNN model uses three-dimensional instead of two-dimensional convolutional and pooling operations. 
 The local CNN models are defined such that they always have the same general structure as the original global CNN but differ in the number of channels of the feature maps, the number of neurons within the fully connected layers, as well as in the number of input nodes. All of the listed layers are proportionally smaller than for the respective global CNN. In particular, each local CNN is trained with input data that correspond to a local part of the original pixel image but has access to all training data points. Consequently, the described approach is a model parallel training method. 
 As output data for each of the local CNNs, which can be trained completely in parallel and independently of each other, we obtain a set of $N$ local probability distributions with respect to the $K$ classes, where each of the local probability distributions corresponds to a local decision exclusively based on information extracted from the local subimages. 
 
 With the aim of generating a final, global decision in form of a global probability distribution with respect to the $K$-class classification problem, we subsequently train a DNN that aggregates the local CNN decisions. More precisely, the DNN uses as input data a vector containing the $K\ast N$ local probability values of all $N$ local CNNs. 
 The DNN model is then trained to map this input vector to the correct classification labels of the original input images corresponding to the $K$ classes of the considered image classification problem. 
 In~\cref{fig:dd_cnn}, we show an exemplary visualization of the described CNN-DNN model architecture for a global CNN of VGG3 type~\cite{simonyan:2014:VGGnet}. The definition and training of the local CNNs is based on the decomposition of the input images into $N=4$ subimages and hence, $N=4$ local CNNs are trained in parallel for this case. Additionally, a DNN is trained to obtain the final, global classification. \\

\noindent \textbf{Comparison with computationally less costly alternatives} 
Besides evaluating the training time and accuracy values of our presented CNN-DNN model, we additionally compare its performance in terms of classification accuracy with two computationally less expensive methods to combine the local classifications of the local CNNs into a final global classification. As a first alternative, we consider the computation of an average probability distribution among the outputs of the local CNNs and assign each input with the label that shows the highest average probability. In~\cref{sec:results}, we refer to this variant as \textit{average probability} (avg. prob.).
Second, we additionally consider a simple majority voting, that is, we assign each image with the label that most of the local CNNs assign their respective subimages to. Let us note that this classification is not necessarily unique since two or more classes may exist which share the majority of the votes. In such cases, we additionally consider the probability values for the respective classes and choose the class among the majority candidates with the highest assigned probability value. In~\cref{sec:results}, we refer to this variant as \textit{majority voting} (maj. vot.). \\

\noindent \textbf{Training the CNN-DNN as one model} Even though the main objective in~\cite{KLW:DNN-CNN:2023} is to provide a network architecture that is well-suited for a model parallel training procedure, additionally, we carefully investigate the classification accuracies of the proposed CNN-DNN model to ensure that the enhanced parallelization is not of the cost of drastically reduced classification performance. Hence, in~\cref{sec:results}, we always compare the accuracy of our CNN-DNN model with a global benchmark CNN which has the same structure and architecture as the local CNNs but with proportionally more parameters and which operates on the entire images as input data. Additionally, for the first time, we also compare the CNN-DNN, where the local CNNs are trained in parallel as described above, with a CNN-DNN that is sequentially trained as one coherent model.
That means that we implement the CNN-DNN architecture as shown in~\cref{fig:dd_cnn} as one model using the functional API of TensorFlow and train it within one sequential training loop. For the remainder of this paper, we refer to this approach as \textit{coherent CNN-DNN} (CNN-DNN-coherent).

\subsection{Transfer Learning}
\label{sec:transfer}

To provide a broader performance test of our proposed network architecture, we further use the concept of transfer learning for the CNN-DNN trained as one model. In this case, we first train proportionally smaller CNNs operating on separate subimages as described in~\cref{sec:CNN-DNN} for $150$ epochs and subsequently use the obtained network parameters as initializations for the respective weights and bias values of the coherent CNN-DNN model. The coherent CNN-DNN model with this initialization is then further trained with respect to the global classification labels of the underlying images. Regarding to the loose analogy of the CNN-DNN training approach to DDMs, the concrete implementation of transfer learning based on locally pre-trained smaller networks can also be interpreted as a preconditioning strategy within an iterative solver or optimization method, respectively; see also~\cite{GuCai:2022:dd_transfer} for a closely related approach for a different global neural network architecture. 
In the following, we refer to this approach as \textit{CNN-DNN with transfer learning} (CNN-DNN-transfer).

\section{Experiments}
\label{sec:results}

In this section, we present some experiments with respect to the described approaches and compare the classification accuracies for different image recognition problems. All experiments have been carried out on a workstation with 8 NVIDIA Tesla V100 32GB GPUs using the TensorFlow library.

\subsection{Network architectures and datasets}

To evaluate the performance of the described training strategies from~\cref{sec:train_strat}, we consider two different network architectures and three different image classification datasets. 
First, we test our approach for a CNN with nine blocks of stacks of convolutional layers and a fixed kernel size of $3\times 3$ pixels, in case of two-dimensional image data, or $3\times 3 \times 3$ voxels, for three-dimensional image data, respectively. We refer to this network architecture as VGG9 for the remainder of this paper and refer to~\cite{simonyan:2014:VGGnet} for more implementational details of this network model. 
Second, we apply all training strategies to a residual neural network (ResNet)~\cite{he2016deep} with 20 blocks of convolutional layers where we additionally implement skip connections between each block and its third subsequent block; see also~\cite{he2016deep} for more technical details. We refer to this network architecture as ResNet20.  
All networks are trained using the Adam (Adaptive moments) optimizer~\cite{Kingma:2014:MSO} and the cross-entropy loss function. 

\begin{figure}[t]
\centering
\includegraphics[width=0.3\textwidth]{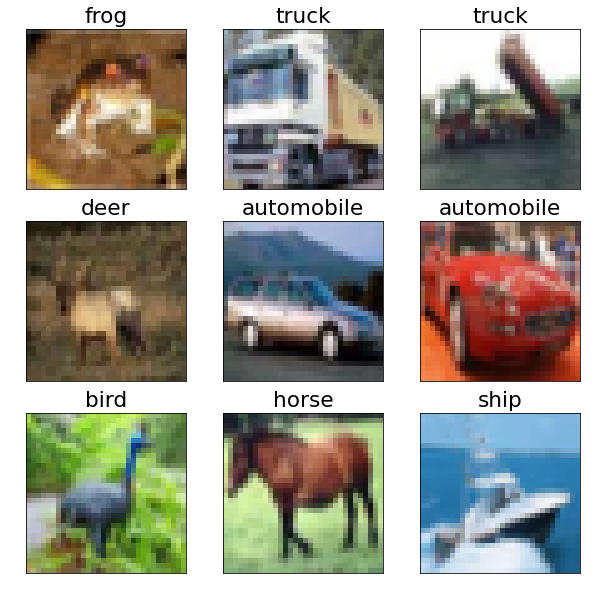}
\hspace{1cm}
\includegraphics[width=0.3\textwidth]{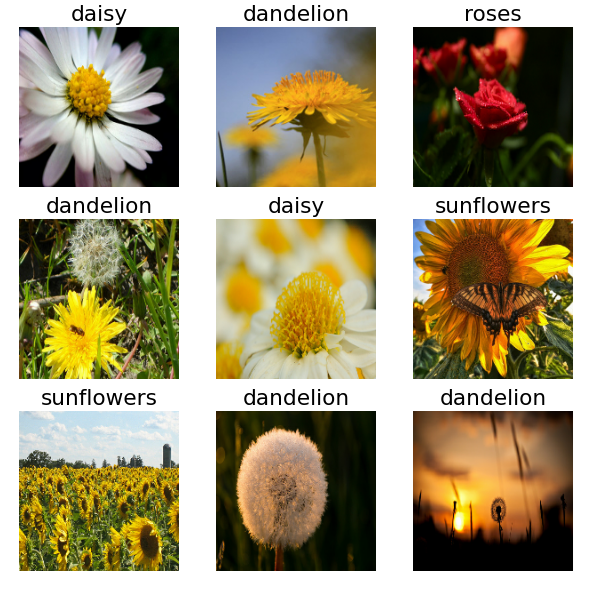}
\caption{\textbf{Left:} Exemplary images of the CIFAR-10 dataset~\cite{Cifar10_TR}. \textbf{Right:} Exemplary images of the TF-Flowers dataset~\cite{tfflowers}.}
\label{fig:cifar10_ex}
\end{figure}

\begin{figure}[t]
\centering
\includegraphics[width=0.5\textwidth]{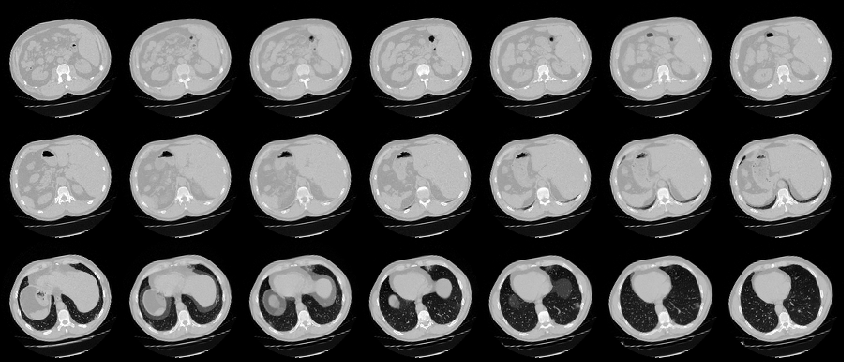}
\caption{Exemplary slices for one chest CT scan taken from the MosMedData dataset~\cite{Chest_CT}.}
\label{fig:CT_lung_ex}
\end{figure}

We test both network models for the CIFAR-10 data~\cite{Cifar10_TR}, the TF-Flowers dataset~\cite{tfflowers}, and a three-dimensional dataset of chest CT scans~\cite{Chest_CT}. 
The CIFAR-10 dataset~\cite{Cifar10_TR} consists of $50\,000$ training and $10\,000$ validation images of $32\times 32$ pixels which are categorized in $K=10$ different classes; see also~\cref{fig:cifar10_ex} (left). Given that these images are relatively small, we only decompose the images into $N=4$ subimages. The TF-Flowers dataset~\cite{tfflowers} consists of $3\,670$ images which we split into $80\%$ training and $20\%$ validation data. All these images have $180\times 180$ pixels and are classified into $K=5$ different classes of flowers; cf. also~\cref{fig:cifar10_ex} (right). 
As the last dataset, we consider the three-dimensional image set of chest CT scans~\cite{Chest_CT} which consists of CT scans with and without signs of COVID-19 related pneumonia, that is, we have $K=2$. For an exemplary visualization of CT slices for one exemplary datapoint, see~\cref{fig:CT_lung_ex}. Each of these CT scans consists of $128\times 128 \times 64$ voxels and hence, here, we train CNN models using three-dimensional filters and pooling layers. 
For all datasets, the DNN model consists of four hidden layers; see~\cite{KLW:DNN-CNN:2023} for more details.

\subsection{Results}

In~\cref{tab:avg_maj}, we compare the classification accuracies for the validation and training data for the CNN-DNN approach for a VGG9 model with a majority voting and an average probability distribution to combine the local CNN classifications into a global decision. As we can observe, for all tested datasets, the CNN-DNN approach results in higher validation accuracies than both, the average probability distribution and the majority voting, for all tested decompositions. This shows that it is not a trivial task to combine the local classifications obtained from the local CNNs into a final, global classification and that it seems to be helpful to train a small DNN to make this evaluation automatically for us. 

When observing the results in~\cref{tab:CNN-DNN_oneModel}, two major observations can be made. First, with respect to the CNN-DNN-coherent model, we see that for the VGG9 model, the coherent model trained in one sequential training loop results in lower validation accuracies than the CNN-DNN model for all three considered datasets and all tested decompositions. However, for the ResNet20 model, the quantitative behavior is reversed, that is, the CNN-DNN-coherent networks result in higher classification accuracies with respect to the validation data. A possible explanation for this could be as follows. The CNN-DNN-coherent model which is implemented as one connected model architecture might have a more complex loss function and thus, loss surface than the locally trained smaller CNNs as well as the relatively small DNN. Hence, optimizing the respective parameters of the VGG9 model all at once might be more difficult than optimizing first the parameters of the local CNNs in parallel and subsequently, the parameters of the DNN. However, when considering the ResNet20 model, the optimization of the CNN-DNN-coherent model might be easier given that the introduction of skip connections usually results in smoother loss surfaces for deep neural networks and enhanced training properties; see also~\cite{he2016deep,li2018visualizing}. This could explain that for the ResNet20 model, the CNN-DNN-coherent shows an improved classification accuracy. A detailed investigation of the resulting loss surfaces and their complexity for the tested models is a potential topic for future research.

Second, when considering the transfer learning strategy, we observe higher classification accuracies for both, the VGG9 network model and the ResNet20 model for all tested datasets compared to the training strategies without transfer learning. Hence, using DDM for means of preconditioning and transfer learning can help to further increase the accuracy of image classification models; cf. also~\cite{GuCai:2022:dd_transfer}. A detailed investigation of the required training times of the transfer learning strategy for our proposed model architecture is a further topic for future research.

\begin{table}[t]
\centering
\caption{\label{tab:avg_maj} Classification accuracies for the validation and training data (in brackets) for the CNN-DNN approach for a  VGG9 model and computationally less costly alternatives to combine the classifications of the local CNNs. In particular, we show the obtained accuracy values for an average probability distribution (avg. prob.) and a majority voting (maj. vot.). }
\scalebox{0.86}{
\begin{tabular}{l||c|c||c}
    {\bf Decomp.} & {\bf avg. prob.} & {\bf maj. vot.} & {\bf CNN-DNN}  \\\hline \hline
    \multicolumn{4}{c}{\bf CIFAR-10}  \\\hline
type A &  0.6745 & 0.6237 & \bf 0.7669 \\
$2\times 2 $, $\delta=0$ &  (0.7081) & (0.6546) & (0.8071)  \\\hline \hline
    \multicolumn{4}{c}{\bf TF-Flowers}  \\\hline
type A &  0.6162 & 0.5974 & \bf 0.6938 \\ 
$2\times 2,$ $\delta=0$  & (0.6498) & (0.6026) & (0.7552) \\\hline
type A  & 0.7565 & 0.7022 & \bf 0.8471 \\ 
$4\times 4,$ $\delta=0$  & (0.7745) & (0.7238) & (0.8593) \\\hline \hline
    \multicolumn{4}{c}{\bf Chest CT scans}  \\\hline
    type A &  0.8038 & 0.7761 & \bf 0.9143 \\
$2\times2\times1$, $\delta=0$  & (0.8279) & (0.7997) & (0.9357) \\\hline
type A &  0.8024 & 0.7453 & \bf 0.8988 \\
$4\times4\times2$, $\delta=0$  & (0.8409) & (0.7999) & (0.9493) \\\hline
\end{tabular}
}
\end{table}

\begin{table}[t]
\centering
\caption{\label{tab:CNN-DNN_oneModel} Classification accuracies for the validation and training data (in brackets) for a global CNN benchmark model (VGG9 or ResNet20), the CNN-DNN approach as introduced in~\cite{KLW:DNN-CNN:2023}, the CNN-DNN model trained as one coherent model (CNN-DNN-coherent), and a coherent CNN-DNN model trained with a transfer learning approach (CNN-DNN-transfer).}
\scalebox{0.86}{
\begin{tabular}{l||c|c|c|c}
    {\bf Decomp.} & {\bf global CNN} & {\bf CNN-DNN} & {\bf CNN-DNN-coherent} & {\bf CNN-DNN-transfer} \\\hline \hline
    \multicolumn{5}{c}{\bf CIFAR-10, VGG9}  \\\hline
type A &  0.7585 & {\bf 0.7999} & 0.7515 & {\bf 0.8462} \\
$2\times 2 $, $\delta=0$ &  (0.8487) & (0.8663) & (0.7902) & (0.8889)  \\\hline
\multicolumn{5}{c}{\bf CIFAR-10, ResNet20}  \\\hline
type A &  0.8622 & 0.8784 & {\bf 0.8998} & {\bf 0.9117} \\
$2\times 2 $, $\delta=0$ &  (0.9343) & (0.9467) & (0.9558) & (0.9664) \\\hline \hline
    \multicolumn{5}{c}{\bf TF-Flowers, VGG9}  \\\hline
type A &  0.7887 & {\bf 0.8154} & 0.7808 & {\bf 0.8378}\\ 
$2\times 2,$ $\delta=0$  & (0.9321) & (0.8827) & (0.8667) & (0.8999)\\\hline
type A  & 0.7887 & {\bf 0.8589} & 0.7676 & {\bf 0.8608} \\ 
$4\times 4,$ $\delta=0$  & (0.9321) & (0.8872) & (0.7995) & (0.8806) \\\hline 
 \multicolumn{5}{c}{\bf TF-Flowers, ResNet20}  \\\hline
 type A  & 0.8227 & 0.8475 & {\bf 0.8776} & {\bf 0.8997} \\ 
$2\times 2,$ $\delta=0$  & (0.9178) & (0.9454) & (0.9603) & (0.9702) \\\hline 
type A  & 0.8227 & 0.8068 & {\bf 0.8406} & {\bf 0.8654} \\ 
$4\times 4,$ $\delta=0$  & (0.9178) & (0.8892) & (0.9002) & (0.9244) \\\hline \hline
    \multicolumn{5}{c}{\bf Chest CT scans, VGG9}  \\\hline
    type A &  0.7667 & {\bf 0.9143} & 0.8889 & {\bf 0.9304} \\
$2\times2\times1$, $\delta=0$  & (0.8214) & (0.9357) & (0.9097) & (0.9577)  \\\hline
type A &  0.7667 & {\bf 0.8988} & 0.8774 & {\bf 0.9025} \\
$4\times4\times2$, $\delta=0$  & (0.8214) & (0.9493) & (0.9305) & (0.9488) \\\hline
\end{tabular}
}
\end{table}

\bibliographystyle{abbrv}
\bibliography{dnn_cnn_dd28}{}

\end{document}